\documentclass[11pt]{article}
\usepackage{nodalida2025}
\usepackage{times}
\usepackage{hyperref}
\usepackage{latexsym}
\usepackage[T1]{fontenc}
\usepackage[utf8]{inputenc}
\usepackage{microtype}
\usepackage{inconsolata}
\usepackage{amsmath}
\usepackage{graphicx}
\usepackage{enumitem}
 \usepackage{xcolor}

\usepackage{listings}
\usepackage{bbm}
\usepackage{amssymb}
\usepackage{lipsum}
\usepackage{tabularx}
\graphicspath{{imgs/}}

\aclfinalcopy 

\title{Danoliteracy of Generative Large Language Models}

\author{
    \textbf{Søren Vejlgaard Holm\textsuperscript{1,2}},
    \textbf{Lars Kai Hansen\textsuperscript{1}},
    \textbf{Martin Carsten Nielsen\textsuperscript{2}}
    \\
    \\
    \textsuperscript{1}Technical University of Denmark, Anker Engelunds Vej 1,  2800 Kongens Lyngby, Denmark,
    \\
    \textsuperscript{2}Alvenir, Applebys Plads 7, 1411 København K, Denmark
    \\
    \small{
        \textbf{Correspondence:} \href{mailto:swiho@ddtu.dk}{swiho@dtu.dk}
    }
}

\begin{document}
\maketitle
\begin{abstract}
The language technology moonshot moment of Generative Large Language Models (GLLMs) was not limited to English:
These models brought a surge of technological applications, investments, and hype to low-resource languages as well.
However, the capabilities of these models in languages such as Danish were, until recently, difficult to verify beyond qualitative demonstrations due to a lack of applicable evaluation corpora.
We present a GLLM benchmark to evaluate \emph{Danoliteracy}, a measure of Danish language and cultural competency across eight diverse scenarios such as Danish citizenship tests and abstractive social media question answering.
This limited-size benchmark was found to produce a robust ranking that correlates to human feedback at $\rho \sim 0.8$ with GPT-4 and Claude Opus models achieving the highest rankings.
Analyzing these model results across scenarios, we find one strong underlying factor explaining $95\%$ of scenario performance variance for GLLMs in Danish, suggesting a $g$ factor of model consistency in language adaptation.
\end{abstract}

\section{Introduction}

\begin{figure}[t]
  \centering
  \includegraphics[width=.85\columnwidth]{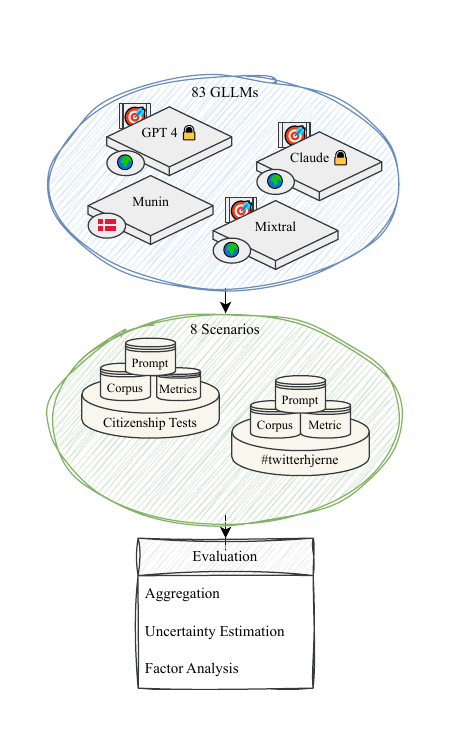}
  \caption{The overall evaluation setup:
  A collection of GLLMs, including closed-source (lock symbol) instruct-tuned (bulls-eye) and multilingual (globe) ones, were evaluated in Danish across diverse use-case scenarios.}
  \label{fig:diagram}
\end{figure}

Benchmarks shape technologies.
By acting as normative guidelines for technology applications, benchmarks imply directions of research and development that ultimately impact users\cite{liang2022holistic}.
GLLMs specifically have emerged as a technology with near-universal impact, including lower-resource languages such as Danish \cite{olsen2023chatgpt}.
If the challenging task of general GLLM evaluation is not extended to low-resource languages, practitioners start from scratch for each model use case, inhibiting practical adoption or possibly resulting in risky, undertested implementations.

We take up the challenge of creating a GLLM evaluation benchmark for Danish, a North Germanic language spoken by 6 million people, primarily in the Nordic country of Denmark.
As depicted in Figure \ref{fig:diagram}, our approach is to create a compilation of small-scale, diverse evaluation scenarios combined into one general benchmark to reveal model Danoliteracy.
By Danoliteracy we refer to the level of GLLM real-world knowledge (RWK), natural language understanding (NLU), and natural language generation (NLG) in Danish.

This paper presents the resulting \emph{Danoliterate Benchmark}, describing evaluation methods, datasets and model results.
We analyze these results with the goals of validating and exploring evaluation methodology.
An important part of this analysis is to investigate the feasibility of such evaluation:
Does this small-scale, language-specific approach achieve a significant ranking of GLLMs?
Even if a non-spurious leaderboard can be discerned from the result, it is not enough to validate the benchmark which might actually show something orthogonal to Danoliteracy.
Thus, as a benchmark validation tool, we additionally present a user survey, collecting the preferences of Danish speakers when interacting with hidden pairs of GLLMs in an arena A/B test setup.

The availability of a suite of meaningful benchmark results allows us to investigate GLLM behavior:
Initially, we explore which specific models are most Danoliterate and how different types of GLLMs compare.
Beyond that, we are particularly interested in capability consistency across tasks:
If a GLLM performs strongly in one Danish use-case scenario, does this performance generalize to other Danish scenarios across different domains and objectives?

We hope so.
If the answer is no, practitioners are without general results to trust, requiring a full model re-evaluation for each downstream use.
However, if capability consistency is present, we should be able to find a single underlying axis that correlates with performance across diverse scenarios.
Such a general dimension of Danoliteracy can be compared to the $g$ factor of general human intelligence \cite{spearman1904general}.
If one significant, main factor is found, it implies a level of stability that can help guide the expectations of practitioners across GLLM implementations in varying and even novel Danish tasks.

The contributions presented in this paper can be summarized as follows:
\begin{itemize}
    \item An open-source benchmark for GLLMs in low-resource languages with an evaluation framework and a live leaderboard site.
    \item The release of a set of novel evaluation datasets for Danish.
    \item Evidence that GPT-4 and Claude Opus models are currently uniquely capable in Danish, outperforming other closed models which in turn overcome open-weights models. 
    \item Evidence suggesting the existence of a Danoliteracy $g$ factor in GLLMs supported by preliminary results from our open-source human feedback study.
\end{itemize}

\section{Related Work}
\subsection{GLLM Evaluation}
The hard task of evaluating free-generation, multitask models has been attempted in many ways.
\citeauthor{liang2022holistic}\ define an empirical approach for revealing model behaviour:
Evaluate each model on a compilation of many scenarios and use-cases of interest, spanning different languages, domains and task categories -- ideally across multiple performance dimensions in addition to raw model capability such as efficiency, bias, and robustness \cite{liang2022holistic}.

This scenario compilation approach has been applied in many ways to GLLMs:
The HELM Lite benchmark presents evaluations of GLLMs on question answering (QA) and translation tasks \cite{liang2023helm}.
Influential benchmarks include the Huggingface OpenLLM Leaderboard \cite{hf2023openllm} and other implementations of the knowledge-based scenarios MMLU \cite{hendrycks2021mmlu} and HellaSwag \cite{zellers2019hellaswag}.

These benchmarks mainly use comparison or similarity algorithms to parse model answers e.g. for finding a chosen option for multiple-choice QA.
Other approaches include applying other GLLMs to grade generations \cite{zheng2023judging} \cite{openai2023evals} or using human feedback \cite{chiang2024chatbot}.

\subsection{Low-resource NLP Evaluation}
Most broadly reported GLLM evaluations are only or primarily performed on examples in English.
Approaches to evaluate lesser-resourced languages include both attempts to compile massively multilingual benchmarks either by automatic translation \cite{lai-etal-2023-okapi} or dataset curation \cite{ahuja-etal-2023-mega}.

Other approaches focus on one language exclusively in attempts to evaluate GLLM language performance beyond surface-level lexical or syntactical literacy.
Using this method, practitioners can align scenario domain, cultural content, and real-world facts with the setting of the language, though a lack of relevant data can be problematic \cite{liu2023nlebenchnorglm}.

Specifically in Danish, the comprehensive ScandEval benchmark, which packages scenarios across eight languages divided into NLU and NLG leaderboards, implements evaluation on GLLMs in Danish on eight NLG scenarios with some overlap in dataset sources with this work \cite{nielsen2023scandeval}.

\subsection{GLLM $g$ Factor}
The idea that GLLM performance is strongly correlated across tasks has been noted previously by for example \citeauthor{ilic023unveiling} who carried out factor analysis on the Open LLM Leaderboard and GLUE, \cite{wang-etal-2018-glue} obtaining results similar to ours \cite{ilic023unveiling}.

\section{Methods}

\subsection{Datasets}\label{sec:datasets}
The eight scenario datasets are divided into three broad categories:
Scenarios testing RWK, scenarios requiring models to perform free NLG and those that imply solving classical NLU tasks.
\paragraph{Real-world Knowledge}
\begin{enumerate}
    \item \textbf{Citizenship Test} is a novel dataset of 605 multiple-choice questions acquired from governmental tests that require applicants for Danish citizenship to demonstrate familiarity with national societal structure, culture, and history \cite{siri2023undervisning}.
    \item \textbf{HyggeSwag} is a novel manual translation\footnote{The text was translated by the authors with each translation being validated by another author completing the inference task.} of 125 HellaSwag \cite{zellers2019hellaswag} ActivityNet \cite{caba2015activitynet} questions testing commonsense natural language inference as a multiple-choice task to pick the only completion consistent with real-world logic.
    \item \textbf{Gym 2000} is a small, novel extraction of 50 literature comprehension multiple-choice questions from the Danish Centre for Reading Research (CRR) aimed at high-schoolers \cite{arnbak2000laesetekster}. 
\end{enumerate}
\paragraph{Free NLG}
\begin{enumerate}
    \item[4] \textbf{\#twitterhjerne} is a novel abstractive question-answering dataset containing 78 anonymized question tweets from the Danish hashtag of that name, translated to \emph{Twitter Brain}, where users ask the social media hive mind for help, input or recommendations. For each question tweet, 2-9 reference answer tweets were extracted making it possible to use the $ \text{score}_\text{o1o}$ metric \eqref{eq:qa_score}.
    \item[5] \textbf{Nordjylland News} is an existing news summarization dataset \cite{kinch2023nordjylland} from which a subset of 300 short news articles with corresponding summaries were used.
\end{enumerate}

\begin{figure}
    \centering
     \includegraphics[width=\columnwidth]{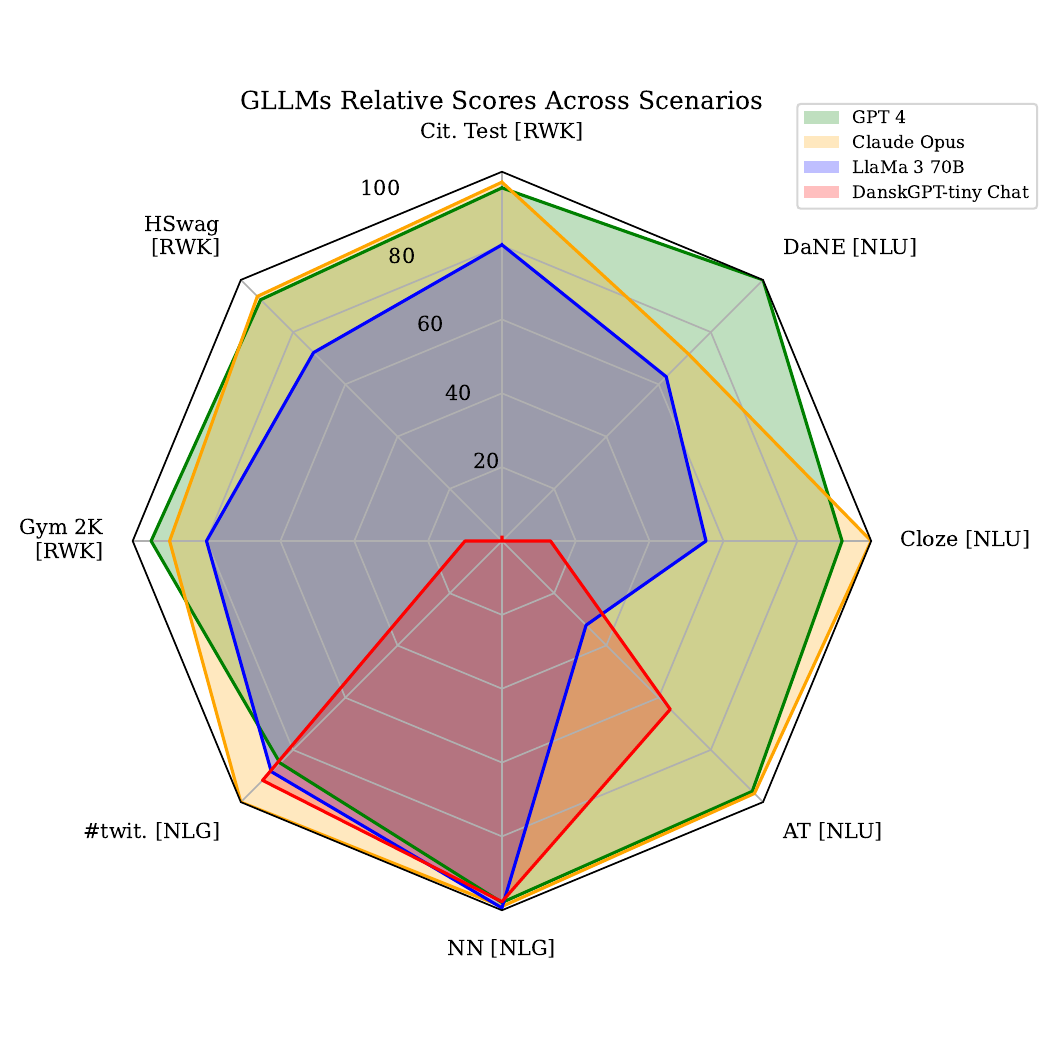}
    \caption{
    Selected model normalized results across the eight scenarios divided into three categories as described in Section \ref{sec:datasets}.
    Claude Opus is overtaken by GPT-4 on the NER task but wins on an NLG task.
    LlaMa 3 70B, the SOTA open-weights model, lags behind on NLU and knowledge-based tasks.
    A Danish-specialized model with only 1.1B parameters, DanskGPT-tiny Chat, benchmarks well in NLG but fails on knowledge and understanding.
    }
    \label{fig:scenarios}
\end{figure}

\begin{figure}
    \centering
     \includegraphics[width=\columnwidth]{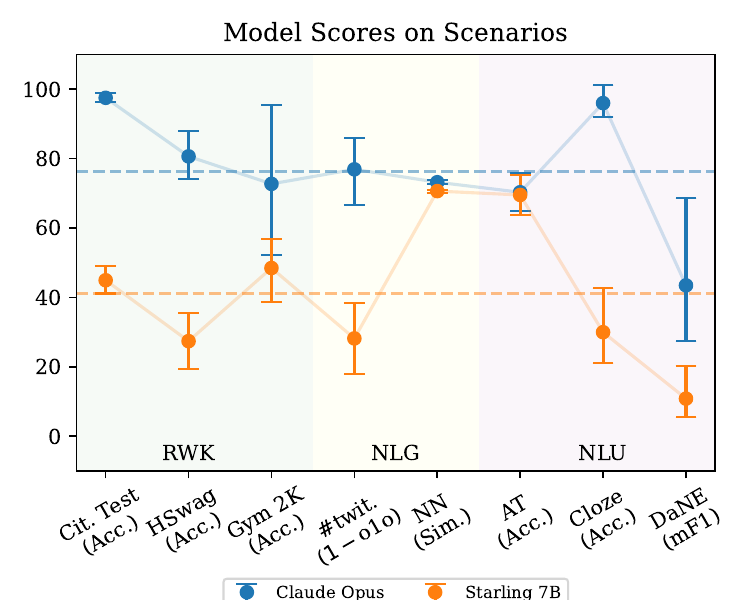}
    \caption{
    The non-normalized metric scores across evaluation scenarios for two models that were judged highly according to human feedback.
    Uncertainties are 95\% confidence intervals according to the bootstrapping procedure and the micro-average is displayed for each model.
k    }
    \label{fig:scenarios2}
\end{figure}

\paragraph{NLU Tasks}
\begin{enumerate}
    \item[6] \textbf{Cloze Self Test} is another small, novel extraction from CRR materials \cite{jensen2015udvikling}, this one containing 33 cloze-style questions evaluated as multiple-choice selection.
    \item[7] \textbf{DaNE} is an existing canonical Danish NER dataset with four entity categories \cite{hvingelby2020dane} from which a subset of 256 examples were used.
    \item[8] \textbf{Angry Tweets} is an existing sentiment classification dataset \cite[Sec. 4]{brogaard2021danlp} with three sentiment categories from which 256 examples were used for multiple-choice prompts.
\end{enumerate}
All datasets are released on the Huggingface Datasets Hub with dataset cards\footnote{Datasets can be found on
\href{https://danoliterate.compute.dtu.dk/Scenarios}{danoliterate.compute.dtu.dk/Scenarios}
} except for the two small datasets extracted from CRR which require practitioners to re-run data collection for personal use.
More details on dataset licensing and collection as well as data examples can be found in Appendix \ref{sec:data_details}.

\subsection{Evaluation}\label{sec:eval_methods}
Each evaluation scenario consisted of a dataset, a prompt template, and a chosen metric.

Most of the available datasets allowed primarily for testing discriminative RWK and NLU of the GLLM by requiring it to select between multiple-choice answers.
For these multiple-choice scenarios, frequency of generating the correct option number was reported as model accuracy.

Two metrics were used for NLG.
First, summarization was implemented using a similarity score between model summary and a reference summary $\operatorname{s}{\left(\mathcal T_{\mathcal M}, \mathcal T_\text{ref}\right)}$.
Secondly, we implemented abstractive question answering tasks for the specific type of dataset $D$ where each question has not just one correct answer but a corresponding set of reference, human-generated answers.
This was done by scoring GLLMs using the frequency with which generated answers were the odd-one-out, defined by the lowest total similarity to all possible answers $T = \{T_{\mathcal M}\} \cup \{\mathcal T_\text{ref, i}\}_{i=1..k}$ as shown in Eq. \ref{eq:qa_score}.
For similarity scores $s$, the BERT score algorithm \cite{zhang2020bert} based on the DFM Encoder Large \cite{danishfoundationmodels2022encoder} was used.
\begin{equation}\label{eq:qa_score}
    \text{score}_\text{o1o} = 
    \mathbb{P}_D\left[
        \mathcal{T}_{\mathcal{M}} = \underset{{t_1 \in T}}{\operatorname{argmin}}
            \sum_{t_2 \in T} \operatorname{s}{\left(t_1, t_2\right)}
    \right]
\end{equation}
Finally, few-shot named entity recognition (NER) was implemented for GLLMs using 3-shot prompting and the GPT-NER multiple queries idea \cite{wang2023gptner}.
Here,  word-level entity class predictions were aligned and the standard NER micro-average F1 scores were calculated using the SeqEval framework \cite{nakayama2018seqeval}.

Scenarios were operationalized by prompting GLLMs in the scenario language, Danish, and structuring prompts with headers marked with the \# character as in Markdown.
In order to use the same prompts for instruct-tuned and base GLLMs, prompts started with the instruction and ended with a text leading towards an answer in the continuation as shown in Figure \ref{fig:prompt_ex}.
Prompting and metric implementation details are covered in Appendix \ref{sec:evaluation}.
\begin{figure}[h]
    \centering
    \begin{lstlisting}[language=TeX]
Write a one-sentence summary of the text.
# TEXT
Lorem ipsum dolor sit amet ...
# SUMMARY
A summary of the text could be:
    \end{lstlisting}
    \caption{
        The general prompting approach translated to English.
    }
    \label{fig:prompt_ex}
\end{figure}\noindent
\begin{figure}[t]
  \centering
  \includegraphics[width=\columnwidth]{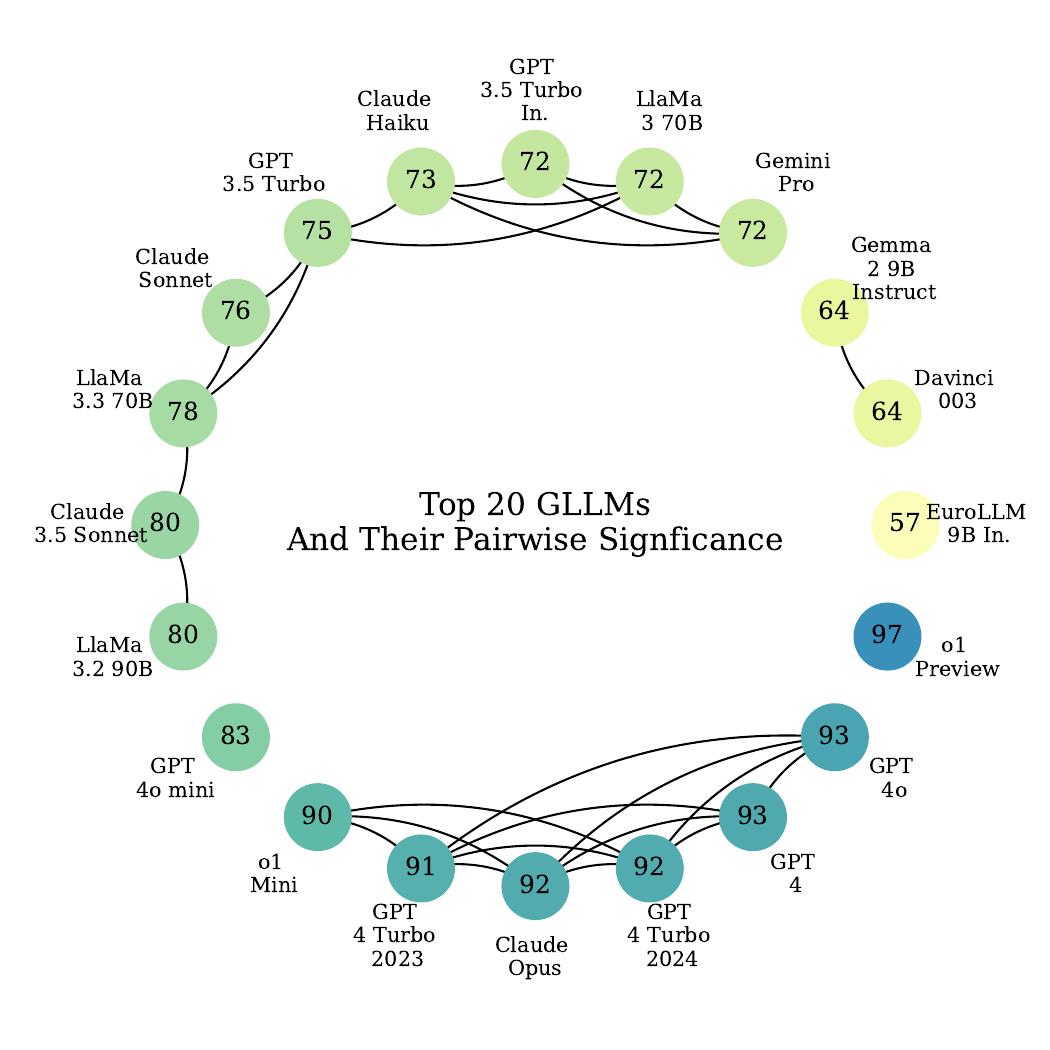}
  \caption{
    Model Danoliteracy Index across all scenarios for top performers.
    Two model nodes are connected iff the bootstrapping procedure could not reveal significant benchmark performance difference at $\alpha=0.05$.
    Together with the special o1 model, Claude Opus and the GPT 4 family models are consistent winners.
  }
  \label{fig:result-significance}
\end{figure}

\subsection{Models}
Both local prediction of open-weights models and API access to externally hosted GLLMs were implemented.
54 autoregressive, decoder language models trained for general text generation were included.
Models were tested if we saw any reason to suspect a degree of Danoliteracy, thus including multilingual models with possibly small amounts of Danish training data as well as other Mainland Scandinavian monolingual models but excluding strictly English-only models. 
Both base and instruct-tuned models were evaluated.
All model generation was performed with greedy decoding and with a maximum number of generated model tokens of 256.
OpenAI's o1 model was allowed to generate internal tokens freely.
The models run locally ranged in sizes from 124M parameters to 13B parameters, resulting in a total project GPU use of $\sim 100$ hours on a single Nvidia H100.

\subsection{The Danoliterate Framework}
A modular, open-source evaluation framework was implemented in Python, using Huggingface Transformers \cite{wolf2019transformers} and Datasets \cite{lhoest-etal-2021-datasets} as central tools as well as Hydra \cite{yadan2019hydra} and a Weights and Biases-integration \cite{wandb} for structuring experimentation.
This framework, \texttt{danoliterate}, is released on GitHub\footnote{
\href{https://github.com/sorenmulli/danoliterate/}{github.com/sorenmulli/danoliterate}
} under the MIT License.

\begin{figure}
    \centering
    \includegraphics[width=\columnwidth]{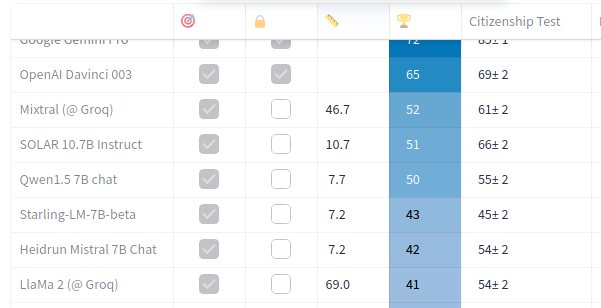}
    \caption{A screenshot from the leaderboard front-end allowing users to explore how model results change with different metric choices as well as inspecting model output examples and reading further details on evaluation scenarios.}
    \label{fig:frontend}
\end{figure}

Furthermore, an interactive site displaying the leaderboard as well as other benchmark results and examples was produced using the Streamlit framework.
See Figure \ref{fig:frontend} for a screenshot of this frontend and Appendix \ref{sec:deps} for versions of software dependencies.

\subsection{Human Feedback}
For a subset of 18 instruct-tuned models, we have set up a parallel study to collect human judgment on model performance.
Volunteers were presented with a anonymized pair of models and were asked to report their preferred model.
This was done based on side-by-side model answers on at least three prompts selected by the volunteer from a pool of 100 prompt examples.
Prompt selection was chosen independently of the Danoliterate Benchmark by creating one Danish prompt for each of 100 popular generative AI use-cases according to \citeauthor{zao-sanders2024genai} \cite{zao-sanders2024genai}.
The study is ongoing: At the time of writing, 477 responses were analyzed.
More details on data collection and analysis can be seen in Appendix \ref{sec:survey}.

\section{Results}
\begin{figure}[t]
  \centering
  \includegraphics[width=\columnwidth]{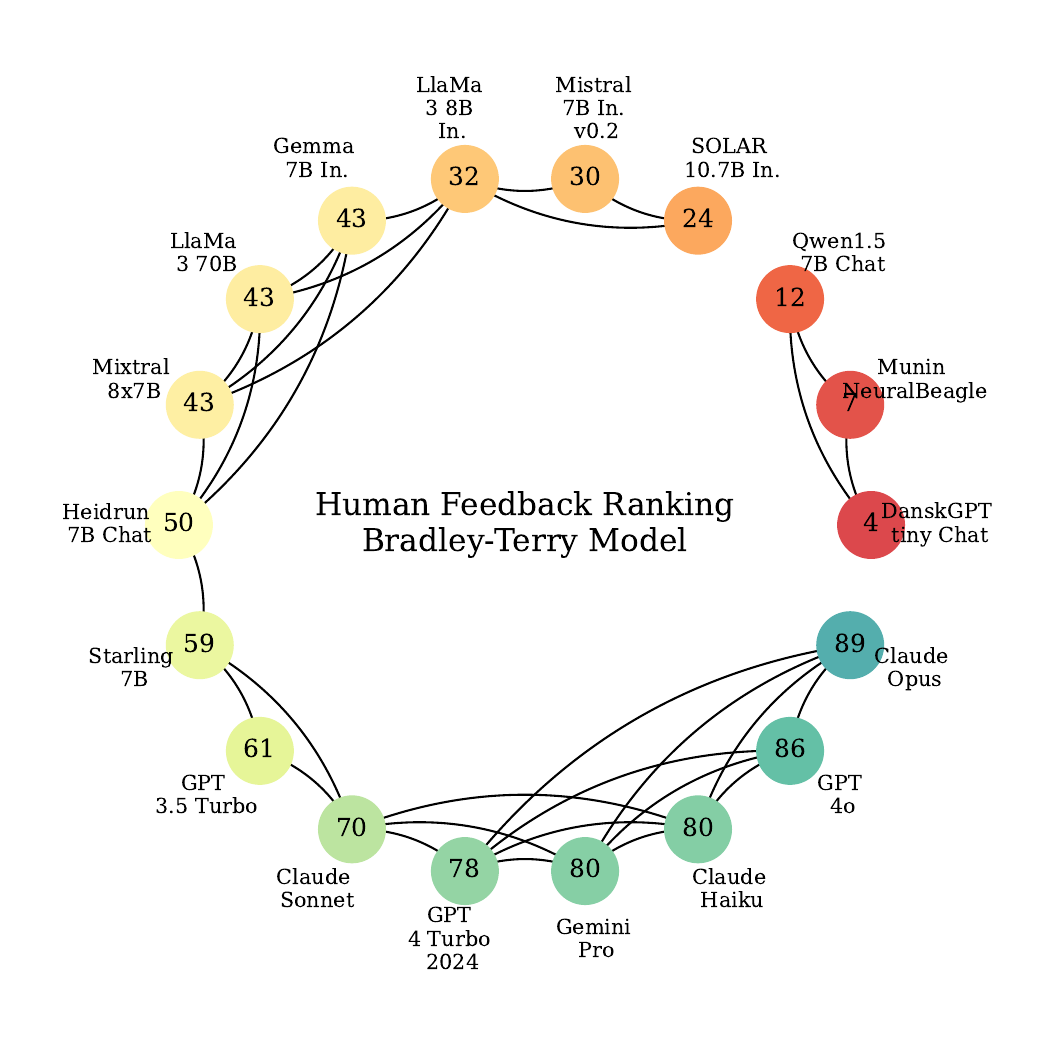}
  \caption{
        Model Danish capabilities based on human feedback.
        Values are normalized Bradley-Terrey coefficients where two models are connected if the coefficients are not significantly different in the ranking model described in Appendix \ref{subsec:ranking}.
  }  \label{fig:human}
\end{figure}

\subsection{Benchmark Feasibility}\label{sec:meaningful}
Benchmarks must have a sufficiently clear signal to be useful.
The ranking in the final leaderboard should be determined by meaningful model differences and influenced minimally by sampling noise.

To quantify benchmark noise, we implemented blocked bootstrapping, resampling all examples with replacement and aggregating all $N=8$ scenario scores for each of the $M=83$ models.
For each $K=10,000$ bootstrap samples, the $M\times N$ model scenario results were aggregated into one overall Danoliteracy Index for each model $d_{\mathcal M}$.
This index was computed by considering one scenario at a time, assigning to the winner index 100 and to the lowest-scoring model index 0 with a linear scaling between the two.
The micro-average across the $N$ scenarios is reported as the resulting Danoliteracy Index of the sample.

The median index is presented as the main leaderboard of this report\footnote{The full $M\times N$ version can be seen at the live leaderboard site:
\href{https://danoliterate.compute.dtu.dk/Leaderboard}{danoliterate.compute.dtu.dk/Leaderboard}.
} with top 20 shown in Figure \ref{fig:result-significance}.
For these, pairwise model index comparisons were performed using the bootstrap samples, correcting $p$ values to control the false discovery rate across $\frac 1 2 M\times(M-1)$ comparisons \cite{BenjaminiHochberg1995}, presenting significant differences at $\alpha=5\%$.

The results show groups of similarly performing models whose Danoliteracy cannot be distinguished.
This increases the lower you go: Many mediocre models, especially non instruct-tuned models, get $d_{\mathcal M}\sim20$: As an example, this small-sample size, curated benchmark cannot reveal a difference between the base models LlaMa 2 7B, $d_{\text{L2}}=20$, and LlaMa 3 8B, $d_{\text{L3}}=23$.

However, robust separation is visible for some models, providing basis for statements like ''The different GPT-4 models benchmark at the same level but clearly perform better than GPT-3 models'' or ''The bigger the Claude 3, the better the performance -- but even the cheap Haiku version performs at GPT-3.5 level'' or ''Small, Danish-specialized models like Heidrun can perform at LlaMa 2 70B level but LlaMa 3 has moved the SOTA for open-weights models in Danish''.
This signal allows us to learn more about reasons for model performance which we explore in the next section.

First, we turn to the important question of validity:
We see a robust benchmark signal resulting in a significant ranking but must question the meaning of the signal.
One superficial indication of a meaningful signal is that, as expected, the ranking correlates significantly with model parameter counts\footnote{Here, only 38 open models with known parameter counts were considered. $\alpha=5\%$ confidence interval: $[0.37;0.78]$.} $\rho\sim0.6$.
However, more importantly: We find that it does correlate with the preliminary results of our Danish human judgement survey.

Ranking human judgement using the Bradley-Terry model as in \cite[Sec. 4]{chiang2024chatbot}, we achieve a ranking shown in Figure \ref{fig:human}.
We observe meaningful differences compared to the Danoliteracy Index:
For example, Claude models are more competitive against GPT 4 and the title as best included open-weights model is taken by Nexusflow Starling \cite{starling2023} from LlaMa 3 70B.
Crucially, however, the general ranking is similar, resulting in a correlation\footnote{$\alpha=5\%$ confidence interval: $[0.6;0.9]$} of $\rho\sim0.8$ with the Danoliteracy Index for these 18 judged models.
We note this as a high value. As a comparison, the Danoliteracy index has a weaker correlation with English benchmarks like HELM Lite and the Open LLM Leaderboard\footnote{This is based on 12 models that overlap between this benchmark and HELM Lite ($\alpha=5\%$ confidence interval: $[-0.2;0.8]$) and the 15 that overlap with the Open LLM Leaderboard ($\alpha=5\%$ confidence interval: $[0.;0.8]$).}, $\rho \sim 0.5$.

Thus, the results from our monolingual scenario compilation approach differ from those from English benchmarks while importantly, showing high correspondence to judgments made by Danish speakers.

\subsection{Model Outcomes}
\begin{figure}[t]
  \centering
  \includegraphics[width=\columnwidth]{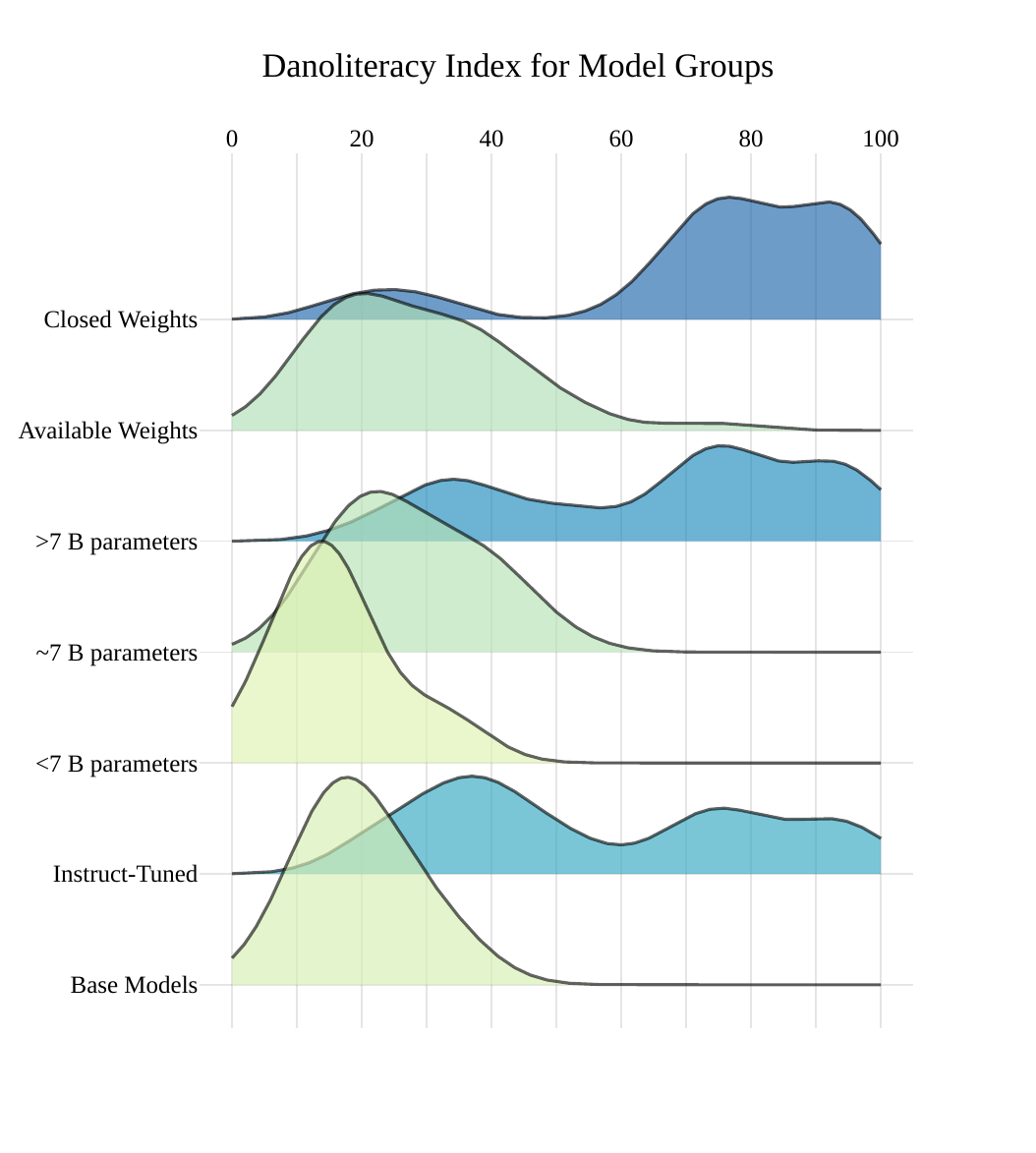}
  \caption{
    Model Danoliteracy Index for groups of models.
    For the models tested in Danish, closed weights dominate open-weights in a remarkably clear way.
    Bigger models are better and on this benchmark, instruct-tuning is necessary to achieve high benchmark scores.
    }
  \label{fig:ridge}
\end{figure}

The leading models are familiar, proprietary top products.
Though LlaMa 3 reaches GPT 3.5 level, Figure \ref{fig:ridge} shows that most models capable in Danish do not have openly available weights.
Furthermore, these top performers are generally also large and instruct-tuned:
Quantitatively, models get about $\sim 0.5$ further Danoliteracy Index points per additional billion parameters and around $\sim 15$ from instruct-tuning\footnote{From fitting a naïve linear model on the results including only models with known parameter counts, see Appendix \ref{sec:outcomes_details}}.

The substantial requirements of dataset and model scale as well as creation of instruct datasets might explain why nationally anchored organizations have not been able to come up with Danish-first models competing with the multilingual behemoths.
Such multilingual models have the advantage of linguistic and factual knowledge enhancement across training languages which also benefit them in the monolingual setting.

\subsection{Capability Dimensionality}
The previous analysis primarily considered the aggregated benchmark results across scenarios.
What is going on at the scenario level?
While different model capability profiles can be seen, exemplified in Figure \ref{fig:scenarios} and Figure \ref{fig:scenarios2}, the main first impression is that model performance at one benchmark scenario strongly predicts performance at other scenarios:
One principal component explains 75\% of the model result variance across the eight scenarios.
\begin{figure}[t]
  \centering
  \includegraphics[width=\columnwidth]{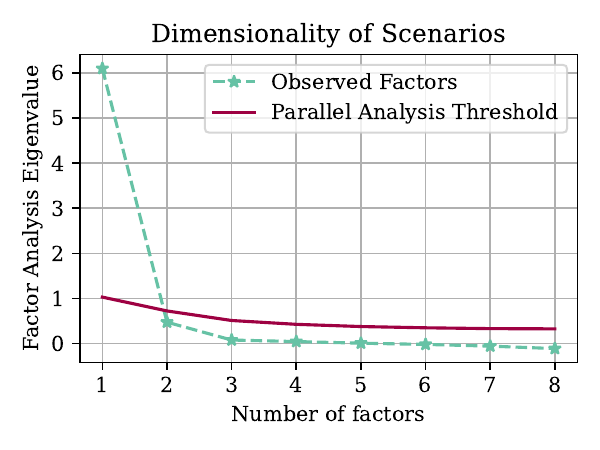}
  \caption{
    Factor Analysis on model results across eight scenarios reveal one underlying dimension of Danoliteracy deemed significant by Horn's Parallel Analysis. 
   }
  \label{fig:pa}
\end{figure}

This finding leads us to the conclusion that a "general factor of Danoliteracy" exists.
We investigate this further using Exploratory Factor Analysis (EFA) on the $M\times N$ scenario result matrix, analyzing the underlying result dimensionality: How many factors are needed to explain the variance induced by model results over the $N$ scenarios?

\begin{figure}[t]
  \centering
  \includegraphics[width=\columnwidth]{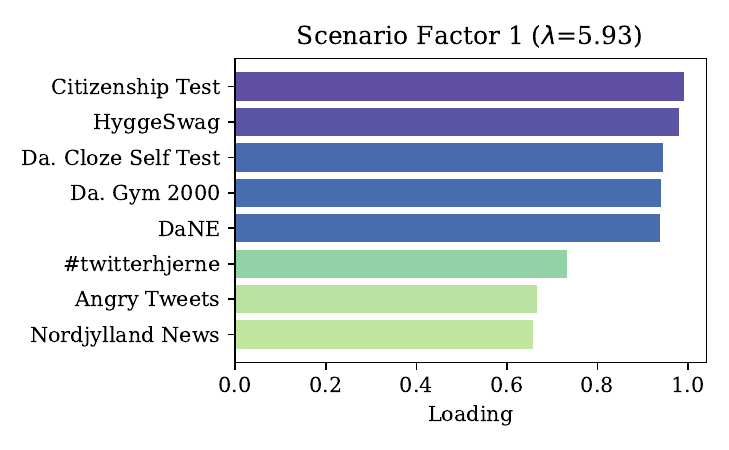}
  \caption{
    The scenarios most contributing to the underlying signal of Danoliteracy:
    The factual multiple-choice evaluation scenario containing Danish citizenship tests most strongly explains benchmark performance.
  }
  \label{fig:scenario-factors}
\end{figure}

This analysis, further detailed in Appendix \ref{sec:fa_checking}, shows a sharp drop in factor eigenvalue when moving from one to two factors as shown in Figure \ref{fig:pa}.
According to both Horn's Parallel Analysis \cite{Horn1965} and the Kaiser Criterion requiring relevant factors to have $\lambda > 1$ \cite{Kaiser1960}, the resulting number of significant factors is 1.
Loadings for this factor is shown in Figure \ref{fig:scenario-factors} suggesting that the RWK scenarios of HyggeSwag and, importantly, the Danish culturally aligned Citizenship Test scenario explain the largest part of the dynamics of the model results.

\begin{table}[h]
    \footnotesize
    \centering
    \begin{tabular}{l|lllll}
         Benchmark & D & $\sigma_\text{F1}$&  $\sigma_\text{F2}$ & F$_\text{K}$  & F$_\text{PA}$    \\
         \hline
    Danoliterate &
    $83\times 8$ & $93\%$ & $7\%$ & 1 & 1 
    \\
    ScandEval Da.     & 
    $199\times 8$ & $88\%$ & $12\%$ & 1 & 2 
    \\
    ScandEval Full     & 
        $199\times 24$ & $77\%$ & $16\% $ & 2 & 2 
    \\
    HELM Lite     & 
            $90\times 10$ & $78\%$ & $19\% $ & 2 & 2 
    \\
    OpenLLM      & 
            $2859\times 6$ & $97\%$ & $3\% $ & 1 & 2 

    \end{tabular}
    \caption{
    How much variance did the first and second factors in EFA explain for the Danoliterate Benchmark as well as the ScandEval benchmark, both full and Danish subset \cite{nielsen2023scandeval}, English benchmarks HELM Lite \cite{liang2022holistic} and OpenLLM \cite{hf2023openllm} as of January 2025.
    Leaderboard dimensionality, model count $\times$ scenario count, is presented along with suggested significant factor count by the Kaiser criterion and Parallel Analysis:
    All benchmarks have an important first component.
    }
    \label{tab:other_fa}
\end{table}
GLLM capability being consistent across different tasks is not just suggested by this benchmark:
Carrying out the same analysis for Scandinavian and English scenario compilations shows an underlying benchmark dimensionality much lower than scenario count, with significant factor count close to 1 as seen in Table \ref{tab:other_fa}.

\section{Conclusions}
Based on the ability to robustly discover model groupings at different Danish capability levels and correlate these rankings with human feedback, we conclude that a scenario compilation approach can meaningfully reveal GLLM capabilities.
We show that, in Danish, open-weights GLLMs currently lag behind large, closed, multilingual, instruct-tuned models, such as GPT-4 and Claude Opus.

For our evaluation setup, we observe one underlying factor in model capability across the diverse test scenarios.
This observation is supported by similar structures in other Danish, English and multilingual scenario compilations which we consider a positive result for low-resource evaluation:
By using curated and language-specific scenarios, the general landscape of GLLM capabilities for a given low-resource language can be meaningfully inferred even if resources limit the scale.

\section{Concerns of the Ethical Impacts}
This work releases a benchmark and leaderboard with the hope of a positive outcome of increased understanding of potentials and limitations of GLLMs in Danish. 
However, we note some risks in the use of such leaderboards.

The results presented here only focus on model capability but, on the leaderboard site, versions of other important dimensions for model applicability are presented; such as model efficiency, model likelihood calibration and model generated output toxicity.
However, these are presented with a disclaimer as preliminary results and our work on other crucial dimensions such as GLLM performance fairness across gender and nationalities or robustness to input noise have not been released due to limitations to current datasets to robustly carry out these analyses.

There is an increased risk of bias, fairness and toxicity violations in low-resource languages to which models are less tuned.
Problematically, when the evaluation situation is also low-resource, these risks might be undiscovered for practitioners that only focus on a model capability.
Further work is crucially needed but for now, the leaderboard site displays a disclaimer against blindly trusting that high benchmark numbers mean predictable downstream performance or applying GLLMs with unchecked assumptions about robustness, fairness, bias, and toxicity.

\section{Limitations}
The study only focuses on one language, Danish, with limited comparisons to other language results.

The presented benchmark consists of eight specific scenarios:
Although we find high correlation between scenario results, all our statements about model performance on Danish in general are evidently biased by the scenario selection.
A similar statement can be made about prompt and metric design decisions though these seem robust in ablation studies in Appendix \ref{sec:evaluation}.

We stress the importance of the uncertainty quantification for this benchmark where all scenarios are small-scale, $n<1,000$:
The bootstrap analysis revealed some model result differences, such as Mixtral 8x7B ($d_{\mathcal M}=54$) and Qwen1.5 7B Chat ($d_{\mathcal M}=50$), are not significant at the desired level and might be spurious.
Other differences such as that between GPT-4 and Claude Opus might be obscured by the important Citizenship Test scenario, where these models achieve close to $100\%$ accuracy (Figure \ref{fig:scenarios2}), being saturated by SOTA GLLMs.
Though most other scenarios still show far-from-perfect accuracy, more difficult scenarios are needed to accommodate future developments.

As all evaluation data is publicly available, unintentional or malignant dataset contamination is possible.
This issue requires attention but might, in the short-term, be less of a risk for low-resource language evaluation with smaller and less widely published corpora.

\section*{Acknowledgments}
We would like to thank all the anonymous reviewers for the insightful and helpful comments.
This work was supported by the Pioneer Centre for AI, DNRF grant number P1.

\bibliographystyle{acl_natbib}
\bibliography{anthology,nodalida2025}

\appendix
\section{Evaluation Methodology}\label{sec:evaluation}
\subsection{Prompting}
An example of a prompt following the structure shown in Figure \ref{fig:prompt_ex} is the Citizenship Test example shown in Figure  \ref{fig:prompt_ex2}.
\begin{figure}[h]
    \centering
    \begin{lstlisting}[language=TeX]
Svar kun med tallet for den rigtige mulighed.
# SPØRGSMÅL
Hvilket af følgende lande har flest indvandrere og efterkommere i Danmark oprindelse i?
# SVARMULIGHEDER
1. Pakistan
2. Iran
3. Tyrkiet
# SVAR
Svaret er mulighed nummer
    \end{lstlisting}
    \caption{
        Prompting of an example from the Citizenship Test scenario.
        Translated, this question prompt reads: \emph{Answer only with the number corresponding to the correct answer. \# QUESTION From which of the following countries do the highest number of immigrants and descendants in Danmark have their roots? \# OPTIONS 1. Pakistan 2. Iran 3. Turkey \# ANSWER The answer is option number.}
    }
    \label{fig:prompt_ex2}
\end{figure}\noindent
\paragraph{Alternatives}
Some prompt alternatives to the approach shown in Figures \ref{fig:prompt_ex} and \ref{fig:prompt_ex2} were run for a subset of GLLMs:
For the Citizenship Test scenario, Table \ref{tab:ct_alt}, results were similar when not presenting options to the models, instead parsing their output choice by selecting the option with highest similarity to their generation.
The same table suggests that changing the Citizenship Test scenario to a simpler prompt without markdown headers lowered results minimally..

For the summarization task, Nordjylland News, an alternative prompt with longer and more detailed instructions had no effect on model results shown in Table \ref{tab:nn_alt}.

Translating all instruction text in the prompt format while keeping data content in Danish maintained or improved non-Danish model results, Table \ref{tab:g2_alt} on the Gym 2000 scenario.
\begin{table}[h]
    \centering
    \small
    \begin{tabular}{l|lll}
 & Std. & Simple Q & No opt. \\
 \hline
Gemini Pro & \underline{\underline{\underline{$85\pm 1$}}} & \underline{\underline{\underline{$78\pm 1$}}} & \underline{\underline{$79\pm 1$}}\\
GPT 3.5 Turbo & \underline{\underline{$82\pm 1$}} & \underline{\underline{$77\pm 1$}}  & \underline{\underline{\underline{$82\pm 1$}}}\\
Mistral 7B Instruct & \underline{$47\pm 2$} & \underline{$50\pm 2$} & \underline{\underline{$49\pm 2$}}\\
Mistral 7B & $45\pm 2$ & $44\pm 2$ & $41\pm 2$ \\
Dano. Mistral 7B & $43\pm 2$ & $45\pm 2$ & \underline{$59\pm 2$} \\
LlaMa 2 7B & $39\pm 2$ & $42\pm 2$ &  $36\pm 2$ \\
Dano. LlaMa 2 7B & $37\pm 2$ & $40\pm 2$ \\
Dummy Baseline & $36\pm 2$ & $36\pm 2$ & $36\pm 2$ 
\end{tabular}

    \caption{
        Alternative prompting and scoring approaches to the Citizenship Test run for a subset of models including a baseline outputting a fixed, random string and Danish-tuned versions of base GLLMs.
        Std. is the prompt version presented in the benchmark, Figure \ref{fig:prompt_ex2}, Simple Q removes the first instruction and the markdown headers, simply presenting the question, the options and the final text.
        No opt. asks the question openly without multiple-choice options, choosing argmax similarity score, as in Section \ref{sec:eval_methods}, as model choice.
        Presented with 95\% Wald confidence interval.
    }
    \label{tab:ct_alt}
\end{table}

\begin{table}[h]
    \centering
        \small
    \begin{tabular}{l|ll}
 & Std. & Detailed \\
 \hline
Gemini Pro & \underline{\underline{\underline{$74\pm 2$}}} & \underline{\underline{\underline{$74\pm 2$}}} \\
GPT 3.5 Turbo & \underline{\underline{$73\pm 2$}} & \underline{\underline{$74\pm 2$}} \\
Mistral 7B Instruct (v0.2) & \underline{$70\pm 2$} & \underline{$71\pm 2$} \\
Mistral 7B & $62\pm 3$ & $58\pm 3$ \\
Dano. Mistral 7B & $57\pm 3$ & $59\pm 3$ \\
Dano. LlaMa 2 7B & $52\pm 3$ & $58\pm 3$ \\
LlaMa 2 7B & $54\pm 3$ & $56\pm 3$ \\
Dummy Baseline & $43\pm 3$ & $43\pm 3$ \\
\end{tabular}
    \caption{
        Impact of Nordjylland News alternative prompting.
    }
    \label{tab:nn_alt}
\end{table}\noindent

\begin{table}[h]
    \centering
    \small
    \begin{tabular}{l|ll}
 & Std. & English \\
 \hline
Gemini Pro & \underline{\underline{\underline{$61\pm 8$}}} & \underline{\underline{\underline{$64\pm 8$}}} \\
GPT 3.5 Turbo & \underline{$45\pm 9$} & \underline{\underline{$52\pm 9$}} \\
Danoliterate Mistral 7B & \underline{\underline{$48\pm 9$}} & $36\pm 8$ \\
Mistral 7B & $39\pm 8$ & \underline{$45\pm 9$} \\
Mistral 7B Instruct (v0.2) & $36\pm 8$ & $42\pm 9$ \\
LlaMa 2 7B & $33\pm 8$ & $33\pm 8$ \\
Danoliterate LlaMa 2 7B & $27\pm 7$ & $30\pm 7$ \\
Dummy Baseline & $21\pm 6$ & $21\pm 6$ \\
\end{tabular}
    \caption{
        How the Gym 2000 results change if the prompt instructions are in English.
        The instruct-tuned models handle this strongly while the Danoliterate Mistral model fails to to perform under English prompting.
    }
    \label{tab:g2_alt}
\end{table}\noindent

\subsection{GLLM Output Parsing}
\paragraph{Multiple-choice}
A numbered option was considered selected if it was the only number generated by the model. In the case of multiple generated option numbers, the most frequent number was chosen.
Maximum generation likelihood-based selection was also implemented and is available for open-weights models on the frontend leaderboard but is not presented here.

\paragraph{GPT-NER}
Following \cite{wang2023gptner}, for each example, the GLLM was prompted four times, once for each entity category in the DaNE dataset \cite{hvingelby2020dane}.
The model was instructed to mark all words belonging to this entity category with \texttt{@}.
These were parsed to one, multi-class prediction, handling overlap by selecting the generation with highest likelihood for models exposing probabilities.
To mitigate small errors resulting in catastrophic results, model output annotated words were aligned using Levenshtein matching to the input example word list \cite{levenshtein1965binary}. 

\subsection{Metrics}
Differences in results for the similarity-based metrics used for \#twitterhjerne and Nordjylland News summarization are presented in Tables \ref{tab:twitter_alt} and \ref{tab:nn_alt2}.
Model rank is minimally changed.

\begin{table}[h]
    \centering
    \small
    \begin{tabularx}{\columnwidth}{l|XXXX}
 & Odd-one-out & Avg. sim. & Min. sim. & Max. sim. \\
 \hline
 GPT 3.5 Turbo & \underline{\underline{\underline{$29\pm 5$}}} & \underline{\underline{\underline{$64\pm 5$}}} & \underline{\underline{\underline{$61\pm 5$}}} & \underline{\underline{\underline{$66\pm 5$}}} \\
 Gemini Pro & \underline{\underline{$31\pm 5$}} & \underline{\underline{$63\pm 5$}} & \underline{\underline{$61\pm 5$}} & \underline{\underline{$66\pm 5$}} \\
 GPT 4 & \underline{$35\pm 5$} & \underline{$63\pm 5$} & \underline{$61\pm 5$} & \underline{$66\pm 5$} \\
SOLAR 10.7B Instruct & $50\pm 6$ & $62\pm 5$ & $60\pm 5$ & $65\pm 5$ \\
LlaMa 2 13B Chat & $60\pm 5$ & $61\pm 5$ & $58\pm 5$ & $63\pm 5$ \\
Mistral 7B Instruct & $64\pm 5$ & $62\pm 5$ & $59\pm 5$ & $64\pm 5$ \\
Dano. Mistral 7B & $96\pm 1$ & $54\pm 6$ & $52\pm 6$ & $57\pm 6$ \\
Dano. LlaMa 2 7B & $97\pm 1$ & $53\pm 6$ & $50\pm 6$ & $55\pm 6$ \\
OpenAI Davinci 002 & $99\pm 0.3$ & $52\pm 6$ & $49\pm 6$ & $54\pm 6$ \\
LlaMa 2 7B & $100$ & $51\pm 6$ & $49\pm 6$ & $53\pm 6$ \\
Mistral 7B & $99\pm 0.3$ & $50\pm 6$ & $48\pm 6$ & $52\pm 6$ \\
Dummy Baseline & $100$ & $44\pm 6$ & $43\pm 6$ & $46\pm 6$
\end{tabularx}
    \caption{Standard version of \#twitterhjerne using the odd-one-out metric \eqref{eq:qa_score} compared to a simpler metric just reporting average similarity score.}
    \label{tab:twitter_alt}
\end{table}

\begin{table}[h]
    \centering
    \small
    \begin{tabularx}{\columnwidth}{l|XXX}
        & BERT similarity & ROUGE-1 & ROUGE-L \\
        \hline
        Gemini Pro & \underline{\underline{\underline{$74\pm 2$}}} & \underline{\underline{\underline{$35\pm 3$}}} & \underline{\underline{\underline{$28\pm 2$}}} \\
        GPT 3.5 Turbo & \underline{\underline{$73\pm 2$}} & \underline{\underline{$32\pm 2$}} & \underline{\underline{$25\pm 2$}} \\
        GPT 4 & \underline{$73\pm 2$} & \underline{$32\pm 2$} & \underline{$23\pm 2$} \\
        SOLAR 10.7B Instruct & $71\pm 2$ & $28\pm 2$ & $20\pm 2$ \\
        Mistral 7B Instruct & $70\pm 2$ & $25\pm 2$ & $18\pm 2$ \\
        LlaMa 2 13B Chat & $69\pm 2$ & $17\pm 2$ & $12\pm 1$ \\
        Mistral 7B & $62\pm 3$ & $16\pm 1$ & $12\pm 1$ \\
        Dano. Mistral 7B & $57\pm 3$ & $11\pm 1$ & $8\pm 1$ \\
        OpenAI Davinci 002 & $55\pm 3$ & $9\pm 1$ & $7\pm 1$ \\
        Dummy Baseline & $43\pm 3$ & $11\pm 1$ & $8\pm 1$ \\
        LlaMa 2 7B & $54\pm 3$ & $6\pm 1$ & $5\pm 1$ \\
        Dano. LlaMa 2 7B & $52\pm 3$ & $5\pm 1$ & $4\pm 0.5$ \\
        \end{tabularx}

    \caption{Nordjylland News summarization results presented with an alternative lemma-based similarity score. The score is computed by lemmatizing text using the SpaCy framework \cite{honnibal2020spacy} and then computing the ROUGE score \cite{lin-2004-rouge}}.
    \label{tab:nn_alt2}
\end{table}

\subsection{Software Dependencies}\label{sec:deps}
The relevant Python packages and their versions are presented in Table \ref{tab:python_versions}. Python version 3.11.8 was used.
\begin{table}[h]
\centering
\small
\begin{tabular}{c|c}
Library & Version \\
\hline
\texttt{google-cloud-aiplatform} & 1.38.1 \\
\texttt{openai} & 0.28.1 \\
\texttt{anthropic} & 0.21.3 \\
\texttt{groq} & 0.4.2 \\
\texttt{pandas} & 1.5.3 \\
\texttt{datasets} & 2.14.5 \\
\texttt{transformers} & 4.36.1 \\
\texttt{torch} & 2.1.1 \\
\texttt{evaluate} & 0.4.0 \\
\texttt{rouge\_score} & 0.1.2 \\
\texttt{bert\_score} & 0.3.13 \\
\texttt{huggingface\_hub} & 0.19.4 \\
\texttt{hydra-core} & 1.3.2
\end{tabular}
\caption{Evaluation framework Python dependencies and used versions.}
\label{tab:python_versions}
\end{table}

\section{Evaluation Corpora Details}\label{sec:data_details}
\subsection{Data Permissions}
\begin{enumerate}
    \item \textbf{Citizenship Test}: All rights reserved ''Styrelsen for International Rekruttering og Integration''. Written permission was given for the data to be re-released as an appendix to Academic work.
    \item \textbf{HyggeSwag}: MIT.
    \item \textbf{Gym 2000}: Unreleased. Written permission was given by CRR for Academic use but not for re-releasing the dataset. 
    \item\textbf{\#twitterhjerne}: CC-By-4.0. 
    \item \textbf{Nordjylland News}: CC-0-1.0.
    \item \textbf{Cloze Self Test}: Unreleased. Written permission was given by CRR for Academic use but not for re-releasing the dataset.  
    \item \textbf{DaNE}: CC-By-Sa-4.0-
    \item \textbf{Angry Tweets}: CC-By-4.0.
\end{enumerate}
\subsection{Data Content}
All novel datasets were manually inspected for offensive content.
Some crime-related and sexual themes were found in Nordjylland News examples but deemed unproblematic.
The \#twitterhjerne dataset was manually anonymized, removing all examples with personally identifiable content.

\subsection{Examples}
Below, one prompted example per evaluation corpus is presented.
\begin{enumerate}[leftmargin=*]
    \item Citizenship Test: See Figure \ref{fig:prompt_ex}.
    \item HyggeSwag
\begin{lstlisting}[language=TeX]
Svar kun med tallet for den rigtige fortsættelse af sætningen
# SÆTNING
En gruppe venner sidder på slæder på toppen af bakken. De to venner
# SVARMULIGHEDER
1. er udstyr kørende ned ad bakken med en udstyrsrem på.
2. presser deres rygge op mod en klippe.
3. skubber en slæde med et reb, da hele bakken er dækket med sne.
4. skubbes ned ad bakken, og de glider til bunden.
# SVAR
Den rigtige fortsættelse er mulighed nummer 
\end{lstlisting}
    \item \#twitterhjerne
\begin{lstlisting}[language=TeX]
Skriv et kort tweet på dansk, der besvarer nedenstående spørgsmål. Svar kun med tweetet.
# TWEET MED SPØRGSMÅL
Sønnen vil gerne lave #pebernødder. De par gange jeg har prøvet det, blev de kun OK. Er der nogen, der kan anbefale en opskrift? #twitterhjerne
# TWEET MED SVAR
Et svar kunne være:
\end{lstlisting}
    \item Gym 2000
\begin{lstlisting}[language=TeX]
"Selv før jeg lærte Max Kelada at kende, var jeg indstillet på ikke at kunne lide ham. Krigen var lige blevet afsluttet, og passagertrafikken på de store oceandampere var livlig. Det var meget vanskeligt at få plads, og man måtte finde sig i at tage, hvad skibsagenterne tilbød én. Man kunne ikke vente at få en kahyt for sig selv, og jeg var temmelig taknemmelig over at få en, hvor der kun var to køjer. Men da jeg erfarede navnet på min medpassager, sank mit humør. Det betød lukkede koøjer, så det ikke ville være muligt at få den mindste smule frisk luft om natten. Det var ubehageligt nok at dele kahyt i fjorten dage med hvem som helst (jeg rejste fra San Francisco til Yokohama), men jeg ville have været mindre bekymret ved tanken, hvis min medpassagers navn havde været Smith eller Brown."

Svar kun med tallet for den rigtige mulighed
# SPØRGSMÅL
Hvorfor var det svært at få en kahyt for sig selv?
# SVARMULIGHEDER
1. Det var moderne at tage på krydstogt.
2. Det var midt i ferieperioden
3. Mange mennesker flyttede til USA
4. Krigen var lige forbi.
# SVAR
Svaret er mulighed nummer 
\end{lstlisting}
    \item Cloze Self Test
\begin{lstlisting}[language=TeX]
"Henrik bladede frem til siderne med boligannoncer. Deres lejlighed var <MASK> for lille til dem, så nu ledte de efter noget større. De ville gerne flytte lidt tættere på kysten. De ledte efter en lille gård , hvor der var plads til at holde et par heste ."
Erstat det maskerede ord i ovenstående tekst (markeret med '<MASK>') med et af følgende ord: indrettet, solgt, annonceret, blevet. Svar *kun* med det rigtige ord:
\end{lstlisting}
\item Nordjylland News
\begin{lstlisting}[language=TeX]
Skriv et kort dansk resumé på én enkelt sætning af følgende tekst.
# TEKST
Manden kom kørende på Sønder Havnevej ved kiosken på havnen i Aalbæk, da han påkørte flere afmærkninger på stedet og fortsatte direkte ind i den bygning, hvor kiosken holder til. Der skete i forbindelse med påkørslen skade på bygningen. Uden for sad en mand, og han blev i lav fart påkørt af bilen ført af 53-årig mand. Den uheldige kiosk-gæst blev kørt til sygehuset med lettere skader. Nordjyllands Politi oplyser, at den 53-årige blev anholdt og sigtet for at køre i spirituspåvirket tilstand. Han er efter endt afhøring løsladt igen.
# RESUMÉ
Et resumé på en sætning er:
\end{lstlisting}
\item Angry Tweets
\begin{lstlisting}[language=TeX]
Vurdér, om sentimentet i følgende tweet er 'positiv', 'neutral' eller 'negativ'. Svar kun med et enkelt ord.
# TWEET
@USER Klæk det æg!
# SENTIMENT:
Sentimentet var
\end{lstlisting}
\item DaNE (prompting for location)
\begin{lstlisting}[language=TeX]
Fuldfør annotering af sidste eksempel i opgaven.
Her er en lingvists arbejde med at annotere entiteter af typen 'lokation'.
# TEKST
Det blev naboens store, sorte hund også, "siger Københavns politidirektør, Poul Eefsen, galgenhumoristisk til B.T. efter et stort smykkekup i hans Holte-villa og en række tilsvarende kup i området.
# ANNOTERING
Det blev naboens store , sorte hund også , " siger @@Københavns## politidirektør , Poul Eefsen , galgenhumoristisk til B.T. efter et stort smykkekup i hans Holte-villa og en række tilsvarende kup i området .
# TEKST
Diskussionen om forklaringen på det "japanske økonomiske mirakel" har især drejet sig om, hvorvidt man kunne nøjes med økonomiske faktorer i sin forklaring, eller om det også er nødvendigt at inddrage særlige kulturelle og historiske forhold for at finde en rimelig forklaring.
# ANNOTERING
Diskussionen om forklaringen på det " japanske økonomiske mirakel " har især drejet sig om , hvorvidt man kunne nøjes med økonomiske faktorer i sin forklaring , eller om det også er nødvendigt at inddrage særlige kulturelle og historiske forhold for at finde en rimelig forklaring .
# TEKST
De lyssky fremmede elementer af enhver art, der har sneget sig til landet, er fjenden.
# ANNOTERING
De lyssky fremmede elementer af enhver art , der har sneget sig til landet , er fjenden .

# TEKST
"Vi tar'en tysker frem, vi tar'en tysker tilbage, vi tar'en tysker frem, åååårrr så ryster vi ham lidt!"
# ANNOTERING
\end{lstlisting}
\end{enumerate}

\section{Survey}\label{sec:survey}
\subsection{Survey Design and Instructions}
For 100 GLLM use-cases divided into six categories \cite{zao-sanders2024genai}, we translated use-cases and categories into Danish and crafted an example prompt in Danish corresponding to that theme.
We saved model answers from 18 models andf used them in the survey to allow interactiveness without requiring infrastructure for true dynamic model responses.

\begin{figure}
    \centering
    \includegraphics[width=\columnwidth]{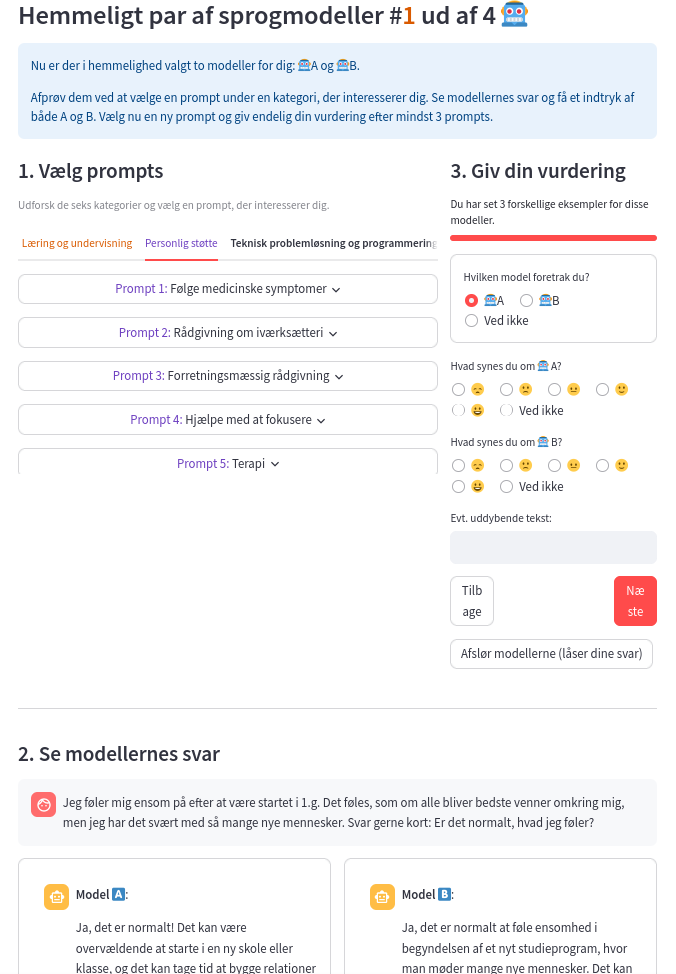}
    \caption{A screenshot of the survey UI presented to users. Under header 1, users select prompts of their interest. Header 2 contains model answers side-by-side and in header 3, volunteers fill in their preference and can get model identities revealed and move on to another pair.}
    \label{fig:survey_screen}
\end{figure}

The survey front-end allowed volunteers to pick between the 100 prompts separated into categories, seeing model outputs from ''Model A'' and ''Model B'' side-by-side, streamed with a delay of 0.1 seconds between each word to simulate model generation.
The volunteer was then instructed to try out at least a total of three prompts before answering.
The answer consists of a question of preference, with optional additional Likert scales for each model and a text field for more details.
The user instruction was in Danish meant \emph{Two models have now been secretly selected for you: Model A and Model B. Test them out by choosing a prompt under a category that interests you. Look at the models' responses and get an impression of both A and B. Now choose a new prompt and please provide your assessment after at least 3 prompts.}\footnote{Da.: \emph{Nu er der i hemmelighed valgt to modeller for dig: Model A og Model B. Afprøv dem ved at vælge en prompt under en kategori, der interesserer dig. Se modellernes svar og få et indtryk af både A og B. Vælg nu en ny prompt og giv endelig din vurdering efter mindst 3 prompts.}}.
See Figure \ref{fig:survey_screen} for an overview of the A/B test user interface.

\subsection{Volunteers}
The survey is openly available online, inviting users to voluntarily try out the A/B tests, filling out their preferences.
The survey was promoted on social media networks and newsletters.
Most of the promotions were made on channels for AI enthusiasts or professionals.
Volunteers were made aware that the data would contrubute to studies into GLLM evaluation in Danish.

\begin{figure}
    \centering
    \includegraphics[width=\columnwidth]{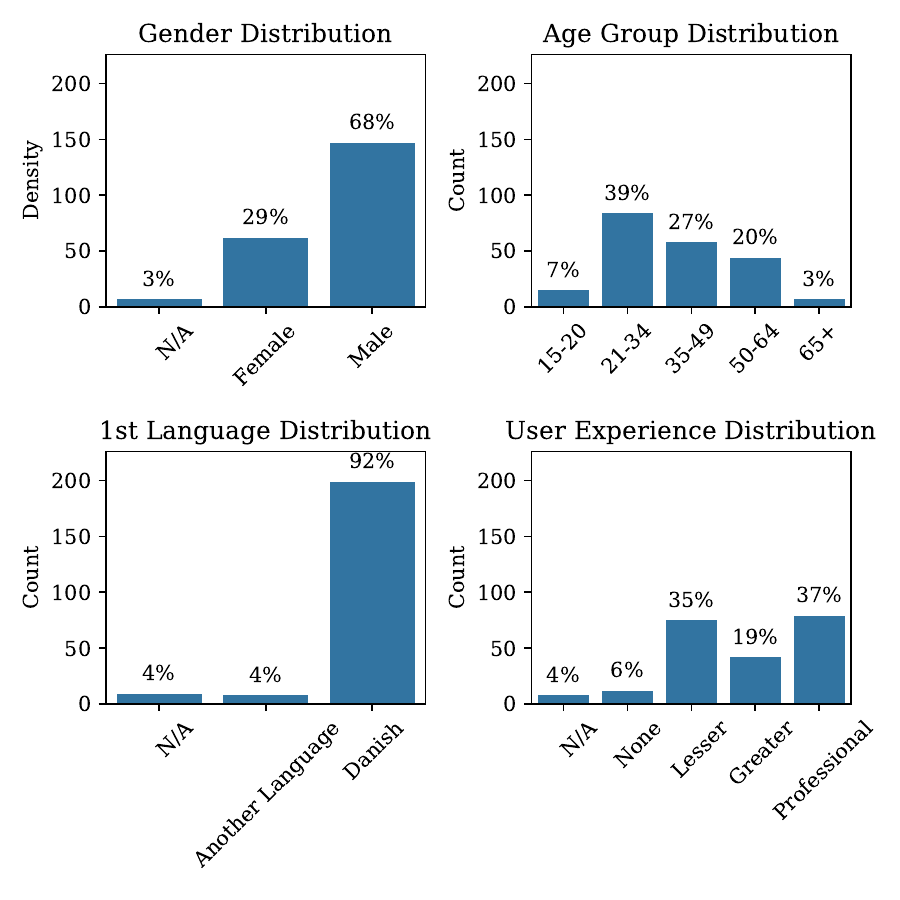}
    \caption{Demographic details filled out by survey volunteers.}
    \label{fig:demo}
\end{figure}

Volunteers could optionally fill in demographic details before carrying out A/B test which is shown in Figure \ref{fig:demo} suggesting a bias towards young, male AI professionals.

Work is ongoing to increase scale and diversity of respondents.

\subsection{Ranking Model}
\label{subsec:ranking}
All 18 models included in the survey were sampled uniformly and the user model preference was used for ranking.
An initial version ranking is the model win frequency presented in Table \ref{tab:winrates}.
As a model for the human preferences, we follow \citeauthor{chiang2024chatbot} to employ the Bradley-Terry model in a non-parametric fashion, using the sandwich robust standard errors \cite[Sec. 4, Sec. 5, Appendix B]{chiang2024chatbot}.
The approach produces a linear model coefficient per model with estimated standard errors.
These can be used for a paired Wilk's test to present significance of differences at $\alpha=0.05$ level.


\begin{table}[h]
    \small
    \centering
    \begin{tabular}{l|lll}
            & Lower & Estimate & Upper \\
            \hline
        Claude Opus & 87\% & 94\% & 100\% \\
        Claude Sonnet & 65\% & 76\% & 89\% \\
        GPT 4o & 65\% & 76\% & 88\% \\
        Gemini Pro & 59\% & 71\% & 85\% \\
        Claude Haiku & 59\% & 71\% & 83\% \\
        GPT 4 Turbo 2024-04-09 & 55\% & 67\% & 80\% \\
        Starling 7B & 53\% & 65\% & 81\% \\
        GPT 3.5 Turbo & 48\% & 61\% & 72\% \\
        Heidrun 7B Chat & 35\% & 48\% & 61\% \\
        Gemma 7B In. & 29\% & 44\% & 60\% \\
        Mixtral 8x7B & 24\% & 40\% & 56\% \\
        LlaMa 3 70B & 24\% & 40\% & 53\% \\
        LlaMa 3 8B In. & 19\% & 32\% & 44\% \\
        SOLAR 10.7B In. & 17\% & 29\% & 41\% \\
        Munin NeuralBeagle & 13\% & 26\% & 39\% \\
        Qwen1.5 7B Chat & 11\% & 22\% & 33\% \\
        Mistral 7B In. v0.2 & 9\% & 20\% & 30\% \\
        DanskGPT-tiny Chat & 6\% & 18\% & 29\%
        \end{tabular}
    \caption{How frequently each model wins their A/B tests with uncertainty estimation to a $95\%$ confidence interval from bootstrapping blocked per volunteer.}
    \label{tab:winrates}
\end{table}

\section{Analysis Methodology Details}
\subsection{Model Outcomes Linear Model}\label{sec:outcomes_details}
The model
\begin{equation}
    d_m = \beta_0 + \beta_1 p_{\mathcal M} + \beta_2 \mathbb I(\mathcal M \in \text{instruct}) + \varepsilon,
\end{equation}
where $\varepsilon \sim \mathcal N(0, \sigma^2)$ and $p_m$ is the number of model parameters in billions, was fitted with results shown in Table \ref{tab:lm}.
\begin{table}[h]
    \centering
    \small
    \begin{tabular}{l|lll}
         Parameter & Value & Std. error & $t$-value \\
         \hline
         $\hat \beta_0$ & 17 2 & 8\\
         $\hat \beta_1$ & 15 & 3 & 5\\
         $\hat \beta_2$ & 0.4 & 0.01 & 4\\
         $\hat \sigma$ & 9
    \end{tabular}
    \caption{
        Linear model fitted with $\hat R^2=0.6$ to the Danoliteracy Index for 38 open-weights models with known parameter counts.
    }
    \label{tab:lm}
\end{table}

\subsection{Factor Analysis}\label{sec:fa_checking}
For the EFA on the $83 \times 8$ scenario results, the Bartlett Sphericity \cite{Bartlett1951} $p$ value is $<2\cdot 10^{-16}$ and the Kaiser-Meyer-Olkin Test \cite{Kaiser1970} yields a variance proportion of $90\%$, both suggesting that the data is usable for EFA.
Fitting an EFA using the Scikit-learn Factor Analysis model yields $\lambda_1=5.9, \lambda_2=0.3$.
Explained factor variance is calculated as eigenvalue proportion of summed eigenvalues,  and the analysis is repeated for scenario results acquired from the open API at \href{https://scandeval.com/danish-nlg/}{scandeval.com/danish-nlg/}, at \href{https://crfm.stanford.edu/helm/lite/v1.3.0/#/leaderboard}{crfm.stanford.edu/helm/lite/v1.3.0/\#/leaderboard}, and using the OpenLLM Leaderboard Scraper GitHub project\footnote{\href{https://github.com/Weyaxi/scrape-open-llm-leaderboard}{github.com/Weyaxi/scrape-open-llm-leaderboard}}.
The datasets updated to most recent versions on January 13th, 2025.

All datasets were subjected Horn's Parallel Analysis \cite{Horn1965} simulating 1000 datasets of same shape but without correlation structure:
This was implemented using the Python package \texttt{horns} \cite{horns2024}.

\end{document}